%% file: LAMDLS.tex
\documentclass[a4paper,UKenglish,cleveref, autoref, thm-restate]{lipics-v2021}
\input{commends}

\title{Latency-Aware 2-Opt Monotonic Local Search for Distributed Constraint Optimization} 

\titlerunning{Latency-Aware 2-Opt Monotonic Local Search for DCOPs} 

\author{Ben Rachmut}{Ben-Gurion University of the Negev}{rachmut@post.bgu.ac.il}{https://orcid.org/0000-0002-3862-9387}{}

\author{Roie Zivan}{Ben-Gurion University of the Negev}{zivanr@bgu.ac.il}{https://orcid.org/0000-0002-1410-8368}{}

\author{William Yeoh}{Washington University in St. Louis}{wyeoh@wustl.edu}{https://orcid.org/0000-0002-2617-870X}{}

\authorrunning{B. Rachmut, R. Zivan, W. Yeoh} 

\Copyright{Ben Rachmut, Roie Zivan and William Yeoh}



\relatedversion{} 

\supplement{}
\supplementdetails[subcategory={}, cite={}, swhid={}]{Software}{https://github.com/benrachmut/CADCOP_CP_2024}

\acknowledgements{This research is partially supported by US-Israel Binational Science Foundation (BSF) grant \#2018081.}

\EventEditors{Paul Shaw}
\EventNoEds{1}
\EventLongTitle{30th International Conference on Principles and Practice of Constraint Programming (CP 2024)}
\EventShortTitle{CP 2024}
\EventAcronym{CP}
\EventYear{2024}
\EventDate{September 2--6, 2024}
\EventLocation{Girona, Spain}
\EventLogo{}
\SeriesVolume{307}
\ArticleNo{9}

\begin{document}

\maketitle
\sloppy


\begin{abstract}
Researchers recently extended Distributed Constraint Optimization Problems (DCOPs) to Communication-Aware DCOPs so that they are applicable in scenarios in which messages can be arbitrarily delayed. Distributed asynchronous local search and inference algorithms designed for CA-DCOPs are less vulnerable to message latency than their counterparts for regular DCOPs. However, unlike local search algorithms for (regular) DCOPs that converge to $k$-opt solutions (with $k > 1$), that is, they converge to solutions that cannot be improved by a group of $k$ agents), local search CA-DCOP algorithms are limited to $1$-opt solutions only.

In this paper, we introduce Latency-Aware Monotonic Distributed Local Search-2 (LAMDLS-2), where agents form pairs and coordinate bilateral assignment replacements. LAMDLS-2 is monotonic, converges to a $2$-opt solution, and is also robust to message latency, making it suitable for CA-DCOPs. Our results indicate that LAMDLS-2 converges faster than MGM-2, a benchmark algorithm, to a similar 2-opt solution, in various message latency scenarios.

\end{abstract}

\section{Introduction}

A promising multi-agent approach for addressing distributed applications, where agents aim to achieve mutual optimization goals, is by modeling them as \emph{Distributed Constraint Optimization Problems} (DCOPs)~\cite{modi:05,petcu:05,fioretto:18}. An illustrative example of such an application is a smart home, where various smart devices must coordinate to create a schedule that optimizes user preferences and satisfies constraints~\cite{Fioretto0P17,rust2022resilient}. In this context, decision-makers are represented as ``agents'' that assign ``values'' to their respective ``variables'', and the objective is to optimize a global objective in a decentralized manner.

DCOPs are NP-hard~\cite{modi:05} and, thus, considerable research effort has been devoted to developing incomplete algorithms for finding good solutions quickly~\cite{yokoo1996distributed,maheswaran:04b,zhang:05,Basharu05,farinelli:08,Smith10,hoang:18,nguyen:19}. Distributed local search algorithms such as \emph{Distributed Stochastic Algorithm} (DSA)~\cite{zhang:05} and \emph{Maximum Gain Message} (MGM)~\cite{maheswaran:04b} are two of the most popular incomplete DCOP algorithms.


Most state-of-the-art local search DCOP algorithms (including DSA and MGM) are synchronous. However, the general setting in which agents operate is inherently \emph{asynchronous}. 
Synchronization is achieved through message exchanges in each iteration of the algorithm, in which an agent receives messages sent by its neighbors in the previous iteration, performs computation, and sends messages to all its neighbors~\cite{zhang:05,Zivan14}. This ensures that at iteration $k$, an agent has access to all information sent to it during iteration $k-1$.
The synchronous design enables the attainment of some desirable properties. For example, MGM agents achieve monotonicity on the quality of the solutions found by modifying their value assignments while ensuring that neighboring agents do not concurrently replace their assignments~\cite{maheswaran:04b}. 

There exists a class of local search DCOP algorithms that guarantee that the solutions found are $k$-opt (i.e.,~they cannot be improved by a group of $k$ agents)~\cite{pearce:07}. MGM is a $1$-opt algorithm and MGM-2 is an extension that is a $2$-opt algorithm. 
Unfortunately, their synchronous designs take advantage of the overly simplistic communication assumptions in the DCOP model, which do not reflect real-world scenarios. Notably, the assumption that all messages arrive instantaneously or with negligible and bounded delays is impractical, given that real-world networks may suffer from delays due to congestion and limited bandwidth. 

To address these limitations, researchers introduced \emph{Distributed Asynchronous Local Optimization} (DALO), an asynchronous $k$-opt algorithm for solving DCOPs~\cite{kiekintveld2010asynchronous}. Unfortunately, its design lacks robustness in scenarios with message delays, restricting its applicability. Specifically, agents try to form groups by asking others to commit to the process they initiate, ensuring an up-to-date local view when computing local optimization. Because neighboring agents attempt to form groups simultaneously, a randomly set local timer is used. Agents can only commit to other groups if a lock request is sent during this timer's duration. However, this design fails when the local timer is not coordinated with the magnitude of message delays, resulting in agents rejecting each other's requests. Additionally, DALO's design does not adequately handle messages not arriving in the order that they were sent. This raises concerns about the algorithm's guaranteed properties under such conditions. 

Recent studies~\cite{rachmut2021latency,rachmut2022communication} explored the performance of local search algorithms for solving DCOPs in the presence of imperfect communication, where messages can be delayed. They demonstrated the significant impact of message latency on the performance of synchronous distributed local search algorithms, especially on property guarantees and convergence rates of MGM. Consequently, a $1$-opt \emph{Latency Aware Monotonic Distributed Local Search} (LAMDLS) algorithm was proposed~\cite{rachmut2022communication}. LAMDLS uses an ordered coloring scheme to prevent neighboring agents from replacing assignments concurrently while preventing agents from waiting for messages as they do in MGM. As a result, LAMDLS demonstrates a quicker convergence rate compared to MGM. 


Building on the success of LAMDLS, we advance the research on distributed algorithms that are robust to message delays by proposing LAMDLS-2, which allows agents to form pairs and coordinate their value assignment selection, while maintaining monotonicity and converging to a $2$-opt solution. LAMDLS-2 enables sequential change of values among paired agents. Agents utilize a unique pairing selection process and an ordering scheme that allows concurrent value modifications for unconstrained pairs. We further discuss a scheme that will allow to generation of a similar monotonic $k$-opt algorithm for any $1 \leq k \leq n$ in future studies. We prove the monotonicity of LAMDLS-2 and its convergence to a $2$-opt solution. Our empirical results indicate that LAMDLS-2 converges significantly faster, in environments with a variety of latency patterns, compared to MGM-2, an existing $2$-opt DCOP algorithm.



\section{Background}
\label{Sec:background}

We present background on Distributed Constraint Optimization Problems (DCOPs), $k$-opt algorithms, including the $2$-opt algorithm MGM-2, Communication-Aware DCOPs (CA-DCOPs), and Latency-Aware Monotonic Distributed Local Search (LAMDLS). 

\subsection{Distributed Constraint Optimization Problems (DCOPs)}
\label{Sec:DCOP_defs} 

A DCOP is a tuple $\langle{\mathcal A},{\mathcal X},{\mathcal D},{\mathcal R}\rangle$, where ${\mathcal A}$ is a finite set of agents $\{ A_1,A_2,\ldots,A_n \}$; ${\mathcal X}$ is a finite set of variables $\{ X_1,X_2,\ldots,X_m \}$, where each variable is held by a single agent (an agent may hold more than one variable); ${\mathcal D}$ is a set of domains $\{ D_1, D_2,\ldots,D_m \}$, where each domain $D_i$ contains the finite set of values that can be assigned to variable $X_i$ and we denote an assignment of value $d \in D_i$ to $X_i$ by an ordered pair $\langle X_i, d \rangle$; and ${\mathcal R}$ is a set of constraints (relations), where each constraint $R_j \in {\mathcal R}$ defines a non-negative {\em cost} for every possible value combination of a set of variables and is of the form $R_j : D_{j_1} \times D_{j_2} \times \ldots \times D_{j_k} \rightarrow \mathbb{R}^+ \cup \{0\}$. A {\em binary constraint} refers to exactly two variables and is of the form $R_{ij} : D_i \times D_j \rightarrow \mathbb{R}^+ \cup \{0\}$.
	
A {\em binary DCOP} is a DCOP in which all constraints are binary. Agents are {\em neighbors} if they are involved in the same constraint. A {\em partial assignment} (PA) is a set of value assignments to variables, in which each variable appears at most once. {\em vars(PA)} is the set of all variables that appear in partial assignment $PA$ (i.e.,~$vars(PA) = \{ X_i \mid \exists d \in D_i \wedge \langle X_i,d \rangle\in PA \}$). A constraint $R_j \in {\mathcal R}$ of the form $R_j : D_{j_1} \times D_{j_2} \times \ldots \times D_{j_k} \rightarrow \mathbb{R}^+ \cup \{0\}$ is {\em applicable} to $PA$ if each of the variables $X_{j_1} , X_{j_2} , \ldots , X_{j_k}$ is included in $vars(PA)$. The {\em cost of a partial assignment} $PA$ is the sum of all applicable constraints to $PA$ over the value assignments in $PA$. A {\em complete assignment} (i.e., {\em solution}) is a partial assignment that includes all variables ($vars(PA)={\mathcal X}$). An {\em optimal solution} is a complete assignment with minimal cost.

For simplicity, we assume that each agent holds exactly one variable (i.e., $n=m$) and we focus on binary DCOPs. These assumptions are common in DCOP literature (e.g.,~\cite{petcu:05,YeohFK10}). 

\subsection{\textit{k}-opt and Region-opt Algorithms} 
\label{subsec:k_opt}

Most local search DCOP algorithms are \emph{synchronous}~\cite{zhang:05,maheswaran:04b,Zivan14}. In MGM, a step (in which agents decide on value replacements) includes two synchronous iterations. First, agents receive their neighbors' updated value assignments and seek improving alternatives for their assignments. Next, agents share their maximal gain from a value replacement. An agent replaces its assignment if its gain exceeds all its neighbors' reported gains. MGM guarantees that agents compute cost reductions using up-to-date information and prevents simultaneous assignment changes by neighbors. This leads to monotonic global cost improvement. MGM also guarantees convergence to a $1$-opt solution.

$k$-opt generalizes the $1$-opt solution concept to any case where $k$ agents cannot improve a solution~\cite{maheswaran:04b,pearce:07}. An algorithm ensuring this must allow all possible coalitions of $k$ agents to seek improving assignments. A well-known algorithm that guarantees the convergence to a $2$-opt solution ($k = 2$) is MGM-2. In MGM-2, agents pair with neighbors to coordinate bilateral assignment replacements. MGM-2's step has five synchronous iterations. In the first three, agents attempt to form pairs, exchange information, and identify the best bilateral gains for these pairs. Unpaired agents select the highest unilateral gain possible. In the remaining two iterations, as in standard MGM, each agent evaluates whether its gain (or the gain of its pair) is larger than the gain of all its neighbors. An agent that is part of a pair, must receive the approval of its partner, that their gain is larger than the gain of the partner's neighbors as well.

A general $k$-opt algorithm was proposed by Pearce an Tambe~\cite{pearce:07} and further generalized to region-optimal algorithms by Vinyals \emph{et al.}~\cite{vinyals:11}. A region is defined by groups of agents that are monitored by the same agent. Commonly, these groups are classified according to two parameters: Their size ($k$) and the distance of the agents from the monitoring agent ($t$). In each step of the algorithm, monitoring agents select a group from their region, aggregate their information, select an alternative assignment, calculate the corresponding gain, and propagate it to the neighbors of all agents in the group. Groups with a larger gain than the gains reported by their neighbors replace their assignments.

\subsection{Communication-Aware DCOPs (CA-DCOPs)}\label{subsec:CA_DCOP}

CA-DCOPs~\cite{rachmut2022communication,zivan2021effect} extend standard DCOPs by using a \emph{Constrained Communication Graph} (CCG) to model the communication latency between pairs of agents. Thus, they can model any pattern of imperfect communication. Specifically, each edge $e$ in the CCG represents the imperfect communication between a pair of agents and is associated with a latency distribution function. 



\subsection{Latency-Aware Monotonic Distributed Local Search (LAMDLS)}

LAMDLS~\cite{rachmut2021latency} is monotonic and $1$-opt (like MGM). By allowing agents to consider value assignment replacements using a partial order, it effectively mitigates the impact of message latency and facilitates faster convergence. To establish the partial order structure it uses the \emph{Distributed Ordered Color Selection} (DOCS) algorithm. DOCS divides the agents into subsets, where agents in each subset have the same color. Colors are ordered (i.e.,~there is a mapping from colors to the natural numbers from $1$ to $NC$, where $NC$ is the number of colors). The neighbors of each agent must hold a different color than its own, and the agent must know which neighbors are ordered before it and which after. During the algorithm execution, each agent keeps track of its neighbors' computation steps, updates them with its selection, and performs the $k$-th iteration when neighbors with a lower color index complete $k$ iterations and those with a higher index complete $k-1$ iterations. LAMDLS demonstrates a faster convergence rate compared to MGM, with the difference becoming more noticeable as the magnitude of message delays increases~\cite{rachmut2022communication}. 



\section{LAMDLS-2}


\emph{Latency-Aware Monotonic Distributed Local Search 2} (LAMDLS-2) is a monotonic algorithm that converges to a $2$-opt solution. $2$-opt algorithms, such as MGM-2, achieve this property by allowing all pairs of agents to make an attempt to improve any assignment that the algorithm traverses, unless it is revised before they get their chance. The main difference in LAMDLS-2 is the method used to generate pairs that will cooperatively suggest an assignment revision. In contrast to MGM-2, where a query response process is used to determine pairs, LAMDLS-2 uses DOCS to find an ordered coloring scheme for determining the pairs. Once DOCS selects an order, the pairs are generated deterministically accordingly, and there are no additional messages required for the pairing process. Thus, message latency has smaller deteriorating effects on this algorithm compared to MGM-2. In order to make sure that all pairs of agents get their chance to improve the current assignment, DOCS is performed iteratively, using random agent indexes. This results in random orderings, which eventually allow all possible pairs to be generated. We present the algorithm in more details below.


LAMDLS-2 is composed of two alternating phases: \emph{Ordering} and \emph{Pair Selection}. Algorithm~\ref{alg:LAMDLS-2} presents the pseudocode performed by an agent $A_i$. In the ordering phase, agents select ordered colors using the DOCS algorithm (\ben{lines~$6$ and $12$}). In the pair selection phase, agents select partners and collaboratively adjust assignments using the pairPhase function (\ben{line~$8$}). The algorithm's input includes the set $N(i)$ that includes $A_i$'s neighbors. 


The algorithm starts with agent $A_i$ randomly selecting $value_i$ for its value assignment~\ben{(line~$1$)}. In addition, $A_i$ maintains a step counter $sc_i$, which is incremented each time $A_i$ selects a value assignment, and a step counter for each of its neighbors in the set $v^{N(i)}$. Entry $v^{N(i)}[j]$ is updated when a value assignment update from a neighbor $A_j$ is received. Both $sc_i$ and entries in $v^{N(i)}$ are initialized to~$1$~\ben{(lines~2-3)}.



\begin{algorithm}[t]
  \caption{LAMDLS-2}
  \label{alg:LAMDLS-2}
  
  \begin{algorithmic}[1]
    \algorithmicinputtext $N(i)$

    \State $value_i \leftarrow$ selectRandomValue()
    \State $sc_i \leftarrow 1$
    \State \textbf{for each } $A_j \in N(i): v^{N(i)}[j] \leftarrow 1$
    \State $docsId_i \leftarrow i$
    \State \textbf{for each } $A_j \in N(i): docsIds^{N(i)}[j] \leftarrow j$

    \State $co_i,co^{N(i)}\leftarrow$ DOCS($i$,$docsIds^{N(i)}$)
    
    \State \textbf{while} stop condition not met:
  
    \State \quad  pairPhase($sc_i$, $v^{N(i)}$,$co_i$,$co^{N(i)}$,$docsIds^{N(i)}$)
    
    \State \quad $docsId_i\leftarrow$ random($0$,$1$)
    \State \quad sendDocsId($N(i)$,$docsId_i$)
    \State \quad $docsIds^{N(i)}\leftarrow$recieveAllDocsIds()

    \State \quad $co_i,co^{N(i)}\leftarrow $DOCS($docsId_i$,$docsIds^{N(i)}$)
        
  \end{algorithmic}
\end{algorithm}

\subsection{Ordering Phase}

In the ordering phase, agents use the DOCS algorithm to select ordered colors, as in LAMDLS~\cite{rachmut2022communication}. Following DOCS, $A_i$ receives its selected color $co_i$, and the colors $co^{N(i)}$ are selected by its neighbors. In contrast to LAMDLS, where agents use their indexes within the DOCS procedure to select colors, in LAMDLS-2 the agents use random values ($docsId_i$). $A_i$ retains the $docsId$'s of its neighbors in the set $docsIds^{N(i)}$. Once $A_i$ has completed the pair selection phase, before re-starting DOCS, it selects a new value for $docsId_i$ and waits for the $docsId$ values of its neighbors to be updated in $docsIds^{N(i)}$ (\ben{lines~9-11}). Hence, each time DOCS operates, it uses different values for $docsId$ and $docsIds^{N(i)}$ and, thus, the probability that it would generate distinct values for $co_i$ and $co^{N(i)}$ is very high. In \ben{line~6}, DOCS is initiated before the pair selection phase. Thus, initial values for the $docsId$s are according to the agents' indexes. The use of randomized $docsId$ values in DOCS results in diverse and randomized ordered color selections in the different steps of the algorithm. 

Algorithm~\ref{alg:docs} details the execution of the DOCS method by some agent $A_i$. At the initiation of the algorithm, $A_i$ holds its own $docsId_i$ and the $docsId$s of its neighbors (in $docsIds^{N(i)}$). When the algorithm terminates $A_i$ holds the color it selected ($co_i$) and the colors of its neighbors ($co^{N(i)}$). The algorithm begins by initializing the variables $co_i$ and $co^{N(i)}$ \benLine{(lines~1-2)}. If the value of $docsId_i$ is the smallest among the values in $docsIds^{N(i)}$, $A_i$ sets the value of $co_i$ to $1$ and sends this information to its neighbors. Afterward, $A_i$ remains idle until it receives updated information about the colors selected by its neighbors \benLine{(line~7)}. The algorithm terminates when $A_i$ becomes aware of the colors of all its neighbors and selects a color for $co_i$ \benLine{(line~6)}. Upon receiving updated information about the colors selected by its neighbors, $A_i$ updates $co^{N(i)}$. Then it checks if it can select a color. If a color was not chosen previously and $A_i$ receives the colors of all its neighbors with smaller indices in $docsIds^{N(i)}$, it selects the color with the smallest number that hasn't been chosen by any of its neighbors and sends this color to its neighbors. This process ensures that eventually, the color selected by each agent is different from the colors selected by its neighbors.
To accelerate the convergence process of LAMDLS-2, agents can select values while they select their colors~\benLine{(line~12)}.

\begin{algorithm}[t]
  \caption{LAMDLS-2 color selection DOCS}
  \label{alg:docs}

  \begin{algorithmic}[1]
    \algorithmicinputtext $docsId_i, docsIds^{N(i)}$
    \algorithmicoutputtext $co_i, co^{N(i)}$

    \State $co_i\leftarrow $ None
    \State \textbf{for each } $A_j \in N(i): co^{N(i)}[j] \leftarrow None$

    \State  \textbf{if} min($docsId_i, docsIds^{N(i)}$) \textbf{then:} 
    \State \quad  $co_i\leftarrow $ 1
    \State \quad  send ($N(i)$,$co_i$,$value_i$)

    \State \textbf{while} not aware of all colors\textbf{:}
    \State \quad \textbf{when} color from $A_j$\textbf{:}  
    \State \quad \quad update ($co_j$, $co^{N(i)}[j]$)
    \State \quad \quad update ($value_j$)

    \State \quad \quad \textbf{if} $co_i$ is None and can select color \textbf{then:} 
    \State \quad \quad \quad $co_i\leftarrow $ selectMinAvilableColor($co^{N(i)}$)
    \State \quad \quad \quad  $value_i\leftarrow $ selectValueUnilaterally($co^{N(i)}$)
    \State \quad \quad \quad send ($N(i)$,$co_i$,$value_i$)

    \State  \textbf{return} $co_i, co^{N(i)}$

  \end{algorithmic}
\end{algorithm}

\subsection{Pair Selection Phase}

Like MGM-2, LAMDLS-2 achieves monotonicity and convergence to a $2$-opt solution by allowing agents to form pairs and select the best mutual assignment, while their neighbors avoid replacing their assignments at the same time. The main difference from MGM-2 is the use of the ordered color scheme by agents to decide when to suggest pairing with their neighbors, which neighbor they should make suggestions to, and whether to accept such suggestions from their neighbors. Agent $A_i$ selects $A_j$ as its partner and shares all relevant information, including its current assignment, the content of its domain, its neighbors, their assignments, and its constraints. Then, when allowed, $A_j$ proceeds to calculate the bilateral value assignments for both $A_i$ and itself and notifies $A_i$ about its updated value assignment. The phase concludes when the agent makes a selection of its value assignment (denoted by $value_i$). If the pairing process is successful, $A_j$ selects the value assignment for both $A_i$ and $A_j$. However, if the pairing process fails (i.e., $A_i$ is not paired with any other agent), $A_i$ can unilaterally select its assignment. Following each selection of a value assignment, there is an update of the agent's step counter ($sc_i$), accompanied by a message sent to its neighbors, which includes $value_i$ and $sc_i$.

Below, we provide a more detailed description of the Pair Selection phase and present its pseudocode in Algorithm~\ref{alg:pair_selection_phase}.
Agent $A_i$ divides its neighbors into two sets, $PC(i)$ and $FC(i)$, based on the input variables $co_i$ and $co^{N(i)}$. $PC(i)$ includes neighbors with color indices smaller than $co_i$, while $FC(i)$ includes neighbors with larger color indices. This division is used to determine the selected neighbor ($sn$) that $A_i$ shares its information with. Agents take into consideration $co_i$, $co^{N(i)}$, $sc_i$, and $v^{N(i)}$ while deciding when to initiate partnerships and how to respond to partnership requests. LAMDLS-2 agents exchange three types of messages during the pair selection phase: 
\begin{itemize}
\item {\bf Value}~(\ben{lines~6-11}): Triggers an update of $v^N{(i)}$, which allows agents to initiate partnerships and reply to them.

\item {\bf Reply}~(\ben{lines~12-13}): Contains the value assignment found by the neighbor the agent paired with.

\item {\bf Offer}~(\ben{lines~14-16}): Contains the relevant information sent when an agent offers a neighbor to form a pair. 
    
\end{itemize}

\begin{algorithm}[t]
  \caption{LAMDLS-2 Pair Selection Phase}
  \label{alg:pair_selection_phase}
  
  \begin{algorithmic}[1]
    \algorithmicinputtext $N(i)$,$sc_i$, $v^{N(i)}$,$co_i$,$co^{N(i)}$,$docsIds^{N(i)}$

    \State $varConsist\leftarrow$ [$sc_i$, $v^{N(i)}$,$co_i$,$co^{N(i)}$]
    
    \State $sn,nInfo\leftarrow None$ 

    \State $sn\leftarrow$offer($varConsist$,$sn$,$docsIds^{N(i)}$)

    \State \textbf{while} phase not completed\textbf{:}
    
    \State \quad \textbf{when} receive message from $A_j$\textbf{:}  
    \State \quad  \quad \textbf{if} message is of type value \textbf{then:} 
        \State \quad \quad \quad update($values^{N(i)}[j]$,$v^{N(i)}[j]$)
         \State \quad \quad \quad \textbf{if} message.$sender$ is $sn$ \textbf{:} 
        \State \quad \quad \quad \quad $value_i\leftarrow$ selectValueUnilaterally()
        \State \quad \quad \quad \textbf{else:}

        \State \quad \quad\quad \quad $sn\leftarrow$offer($varConsist$,$sn$,$docsIds^{N(i)}$)

    \State \quad \quad \textbf{if} message is of type reply \textbf{then:} 
        \State \quad \quad \quad updateValue(message.getValue(i))

    \State \quad \quad \textbf{if} message is of type offer \textbf{then:} 
        \State \quad \quad \quad $nInfo\leftarrow$getOfferInfo(message,$docsIds^{N(i)}$)
        \State \quad \quad \quad reply($varConsist$,$nInfo$)

    \State $sc_i\leftarrow sc_i+1$
     \State  sendLocalInfo($N(i)$,$value_i$,$sc_i$)

  \end{algorithmic}
\end{algorithm}


Upon receiving a \textbf{value} message, $A_i$ updates its local view (line {$7$}) and then considers two scenarios that may be triggered: Either rejecting or initiating an offer. If the sender of the value message is the agent ($sn$) to whom $A_i$ has made an offer in the current phase (lines {$8-9$}), $A_i$ considers the value message as a rejection of its offer. Conversely, if $A_i$ did not initiate an offer during the current phase, a value message reception may prompt an offer initiation due to an update in $v^{N(i)}$, as $A_i$ examines the necessary condition to offer (lines {$10-11$}). 

In the offer function, $A_i$ checks its eligibility to make an offer when the condition $sc_i = sc_j-1$ is met for every $A_j \in PC(i)$. The offer function is activated under two circumstances. The first occurs when a value is received from the neighbor $A_j$. This results in an update of $sc_j$, which might satisfy the condition that will allow $A_i$ to offer. The second is tied to the base case that initiates the phase for agents meeting the condition due to $pc = \emptyset$ (line 3). When the agent decides to make an offer, it selects a neighbor ($sn$) using a deterministic process. The chosen neighbor must meet the following conditions: Its color index is larger by one from the color index of $A_i$ ($co_i +1 = co^{N(i)}[sn]$), and the value of $v^{N(i)}[sn]$ equals $sc_i$. If multiple agents meet these conditions, the neighbor with the smallest value in $docsIds^{N(i)}$ is chosen. If $sn$ is found, $A_i$ sends an \textbf{offer} message containing all relevant information for a bilateral value assignment selection. The function returns $sn$ for future examination of whether the offer was accepted or rejected. If no neighbor satisfies the conditions to qualify as $sn$, $A_i$ unilaterally selects a value assignment and indicates that the phase is completed.  After sending an offer message, $A_i$ enters an idle state, awaiting a reply from $sn$. Upon receiving a \textbf{reply} message, $A_i$ is informed of the offer's acceptance. Subsequently, $A_i$ updates its $value_i$ based on the bilateral decision made by $sn$ (line~$13$).

Upon receiving an \textbf{offer} message, $A_i$ stores the shared information and uses the reply function (line~$15$). $A_i$ has the option to either accept the offer or reject it. $A_i$ can only accept a single offer per step. If $A_i$ accepts the offer, it proceeds to calculate values for itself and its partner using its local information and the information received from its partner and sends a \textbf{reply} message back to it. However, if $A_i$ declines the offer, indicating that it has already formed a bilateral value assignment change with a different agent, it sends a message containing its value to inform the sender that the offer was rejected. If $A_i$ receives multiple offers, it selects as a partner the offering agent with the lowest index in $docsIds^{N(i)}$. Let $PO(i)$ denote the set of agents that sent offers to $A_i$ in the current pair selection phase. An offer can be accepted by $A_i$ if the following condition is met: for each agent $A_j \in PC(i)\backslash PO(i)$, $sc_i=sc_j-1$. Until this condition is met, $A_i$ will remain idle and wait for messages to arrive.

\subsection{Demonstration of LAMDLS-2}\label{appen:demoRun}

\begin{figure}[t!]
    \centering
    \begin{subfigure}[t]{0.4\textwidth}
        \centering
        \includegraphics[scale=0.2]{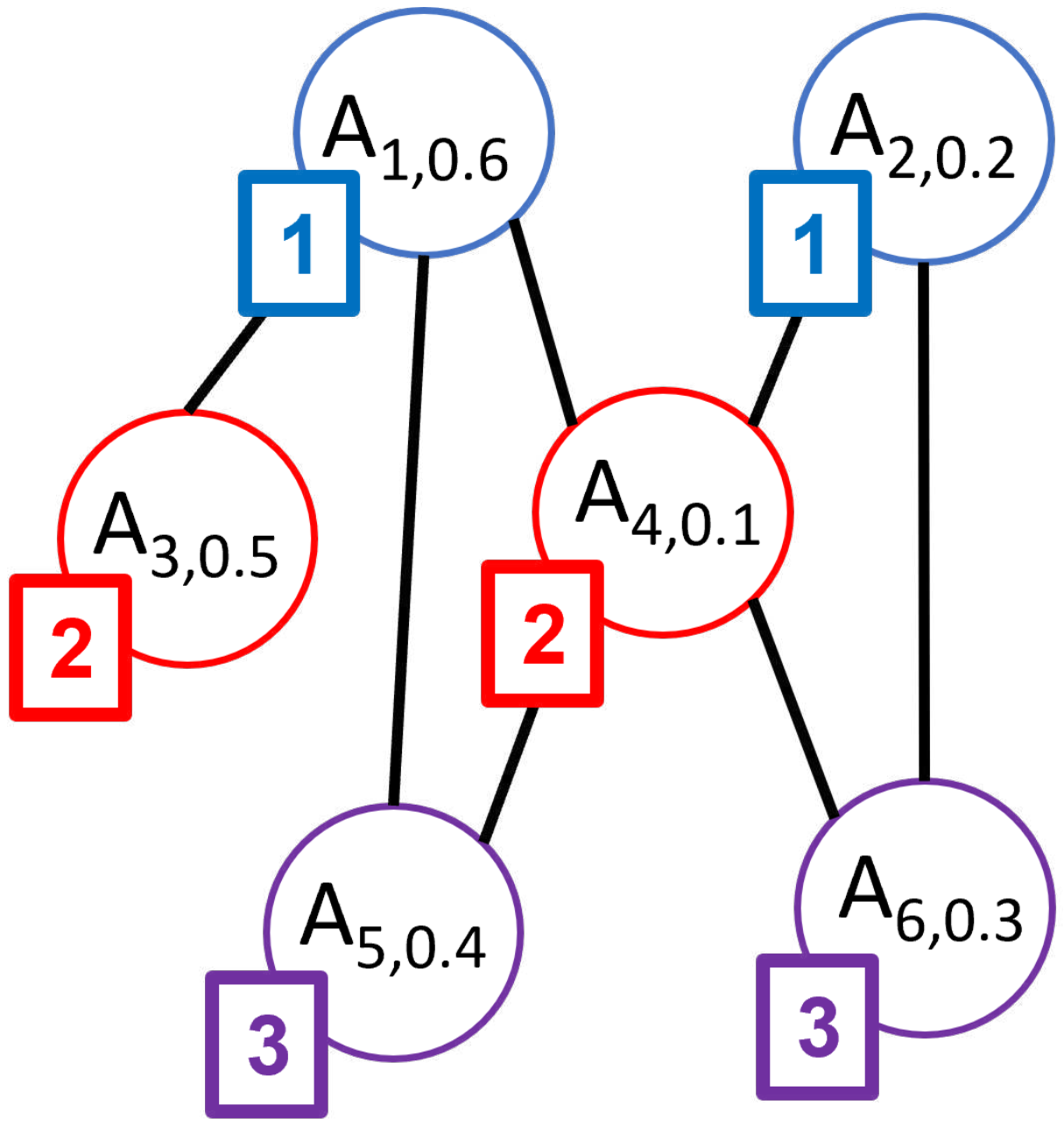}
        \caption{\centering Identity index}
    \end{subfigure}%
    ~ 
    \begin{subfigure}[t]{0.4\textwidth}
        \centering
        \includegraphics[scale=0.2]{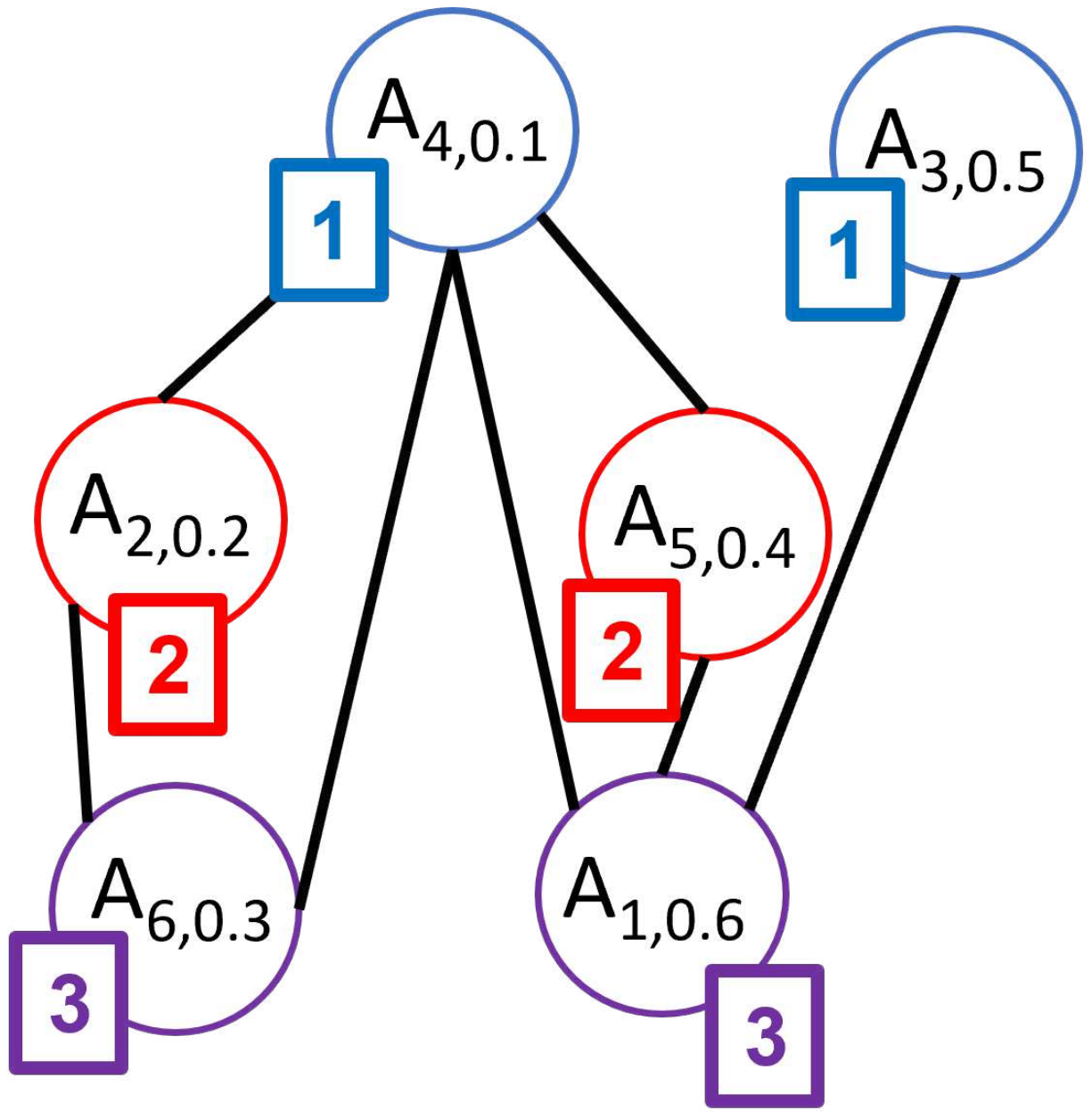}
        \caption{\centering Random index}
    \end{subfigure}
    \caption{\centering Two different numerical graph color partitions.}
    \label{Fig:color_exmps}
\end{figure}


\ben{In the following sub-section, we describe the beginning of a high-level trace of LAMDLS-2, when operating on the constraint graph presented in Figure~\ref{Fig:color_exmps}. In this graph, each node represents an agent, and the corresponding colors (selected using DOCS) of the agents are displayed beneath the nodes. Specifically, each node represents an agent $A_{i,docsId}$, where $i$ is the agent's index and $docsId$ is a randomly assigned value that is drawn before the next step. 
}

\ben{After agents randomly select values for their assignments, each agent initializes its $docsId$. They also set the entries of $docsIds^{N(i)}$ with the identity indices of their respective neighbors, e.g., $A_1$: $docsId_1 = 1$ and $docsIds^{N(1)} = [\langle A_3:3 \rangle, \langle A_4:4 \rangle, \langle A_5:5 \rangle]$
}

\subsubsection*{First Step}

\ben{After initiation, agents proceed to execute DOCS. Figure~\ref{Fig:color_exmps} (a) presents the outcome of the color selection process carried out by DOCS. This process utilizes the values of $docsId$ of the agents, therefore the outcome is dependent on their selection. In the example at hand, agents $A_{1}$ and $A_{2}$ do not have neighbors with smaller indices, so they select the color $1$ (blue) and communicate this information to their neighbors. Among these neighbors, agents $A_{3}$ and $A_{4}$ do not have other neighbors with smaller indices, so they choose the color $2$ (red) and send messages including this information to their neighbors. Finally, agents $A_{5}$ and $A_{6}$ select the color 3 (purple). This completes the color selection phase. }


\ben{When the pair selection phase begins, both $A_1$ and $A_2$, which selected the color $1$, can choose a neighbor and send an offer along with the relevant information. They are eligible because $PC(1) = \emptyset$ and $PC(2) = \emptyset$. $A_1$ must select a neighbor with the smallest $docsId$ color among its neighbors with color $2$. It has two neighbors with color $2$, $A_3$ and $A_4$), and among them, $A_3$ has a smaller $docsId$, thus, it sends the offer to $A_3$. $A_2$ selects $A_4$, since it is its only neighbor with color $2$.}

\ben{Upon receiving an offer, $A_3$ is eligible to respond, given that $A_1$ is its only neighbor. $A_3$ selects values for itself and for $A_1$, updating $sc_3$ to $2$. It then sends a reply to $A_1$, who adjusts its assignment and updates $sc_1$ to $2$, notifying all its neighbors including $A_4$.}

\ben{After receiving an offer from $A_2$, $A_4$ must wait for an update from $A_1$ (which is included in $PC(4)$). Following this update, $A_4$ selects values for itself and for $A_2$, increments $sc_4$ to $2$, responds to $A_2$ and informs its neighbors of the new selected value. Subsequently, $A_2$ updates its value, $sc_2$ becomes $2$, and it informs its neighbors too.}

\ben{At this point, agents with colors $1$ and $2$ have already chosen value assignments. Upon receiving this information, $A_5$ updates $v^{N(5)}=[\langle A_4:2 \rangle,\langle A_5:2 \rangle]$. Thus, when receiving a value message that finalizes the update of $v^{N(5)}$, $A_5$ is eligible to offer, given that $sc_5 = 1$ and $sc_1 = sc_4 = 2$. While attempting to find a suitable partner, $A_5$ will pick a value unilaterally since no agent in $co^{N(5)}$ holds color $4$ (which is one greater than $co_5 = 3$). Similarly, $A_6$ will also independently select its value assignment. This finalizes the second phase of the first step.}

\subsubsection*{Second Step}

\ben{At the beginning of the second step, agents select random $docsId$s and send messages that inform their neighbors of their selection.}

Next, agents execute DOCS using the random $docsId$s selected and generate the color selection that is depicted in Figure~\ref{Fig:color_exmps} (b), as described next: Agents $A_{4}$ with $docsId_4 = 0.1$ and $A_{3}$ with $docsId_3 = 0.5$ do not have neighbors with smaller $docsId$ values, leading them to select color $1$ (blue) and communicate this decision to their neighbors. Agents $A_{2}$ with $docsId_2 = 0.2$ and $A_{5}$ with $docsId_5 = 0.4$ can then select the color $2$ (red) and convey it to their neighbors. Eventually, agents $A_{6}$ with $docsId_6 = 0.3$ and $A_{1}$ with $docsId_1 = 0.6$ select color $3$ (purple) and inform their neighbors.

In the pair selection phase, agent $A_4$ selects $A_2$ as its partner and forwards an offer (since $docsId_4 < docsId_5$, i.e., $0.2<0.4$). Agent $A_3$ changes its value independently, as its only neighbor $A_1$ has color $3$. Upon receiving a value message from $A_4$, $A_5$ can send an offer to $A_1$. After $A_1$ receives a value update from $A_4$, it can respond to $A_5$. Notably, in the previous step, the pair $A_5$ and $A_1$ did not form a partnership. When $A_6$ receives value messages from $A_2$ and $A_4$ ($PC(6) = \{A_2, A_4\}$), it attempts to select a neighbor. Failing to do so (no neighbors in $FC(6)$), it selects a value on its own.

\subsection{Theoretical Properties}

We now prove that LAMDLS-2 is monotonic and convergence to a $2$-opt solution. Our monotonicity proof stems from previous studies that proved the monotonicity of MGM, MGM-2, and LAMDLS~\cite{Maheswaran04a,rachmut2022communication} based on the fact that, in DCOP algorithms, when a single agent or a pair of agents improve their local state, while their neighbors remain idle, the global cost improves as well. Thus, it remains to show that when an agent or a pair of agents improve their local state in LAMDLS-2, their neighbors are idle until the messages regarding the assignment replacements that were performed by the agent or pair of agents arrive.

\begin{lemma}
In a DCOP (with symmetric constraints), when an agent $A_i$ is the only agent replaces its assignment, while none of its neighbors ($NC(i)$) replace their assignments, and this replacement results in a local gain, it also results in an improvement of the global cost.
\end{lemma}

{\bf Proof:} 
Denote the global cost before $A_i$'s assignment replacement by $gc$ and the local gain following $A_i$'s assignment replacement by $LR_i$. Since the problem is symmetric, the sum of local gains of $A_i$'s neighbors is also equal to $LR_i$. Since we assumed that $LR_i > 0$, $gc > gc - 2LR_i$. 
 $\hfill \square$

\begin{lemma}
When some agent $A_i$ initiates a partnership offer, all agents in $N(i)$ that do not partner with $A_i$ avoid replacing their assignments until $A_i$ completes its assignment replacement.
\label{Lem:N(i)}
\end{lemma}

{\bf Proof:} 
For $A_i$ to be active, $sc_i$ must be equal to $k$ (i.e., it has not been incremented since the color selection phase) and, for each agent $A_{i'} \in PC(i)$, $sc_{i'} = k+1$. Thus, when $A_i$ sends an offer, all agents in $PC(i)$ have already incremented their step counters. In addition, for each agent $A_{j'} \in FC(i)$ (i.e., $A_i \in PC(j')$), until $sc_i$ is incremented, $A_{j'}$ cannot send an offer or replace its assignment. 
$\hfill \square$

\begin{lemma}
When agent $A_i$ initiates a partnership offer to $A_j$, agents in $N(j)$ do not replace their assignments until $A_j$ completes its assignment replacement.
\label{Lem:N(j)}
\end{lemma}

{\bf Proof:} 
Agents in $FC(j)$ cannot offer or reply to an offer until $sc_j$ is incremented. On the other hand, for the agents in $PC(j)$, there are two cases:
\begin{itemize}

\item $A_{i'} \in PC(j) (i \neq i')$ \textbf{did not offer to} $A_j$. Then, $A_j$ will not reply and replace assignments until $sc_{i'}$ is incremented, which can happen only after $A_{i'}$ replaces its assignment. Thus, it cannot happen concurrently with the assignment replacement of $A_j$.

\item $A_{i'} \in PC(j) (i \neq i')$ \textbf{did offer to} $A_j$. Then, either $A_j$ pairs with it, or it sends a rejection reply only after it completed the assignment replacement. Thus, they do not replace assignments concurrently. $\hfill \square$
\end{itemize}

\begin{proposition}
\label{Prop:mono}

LAMDLS-2 is monotonic (i.e., each assignment replacement improves the global cost of the complete assignment held by the agents). 
\end{proposition}

{\bf Proof:}
Follows immediately from Lemma~\ref{Lem:N(i)} and Lemma~\ref{Lem:N(j)}. While agents replace their value assignments, none of their neighbors can replace their assignments. 
$\hfill \square$

\begin{proposition}
At each pair selection phase, every agent that receives an offer will reply (positively to one of the offering agents and negatively to the rest). 
\label{Prop:cont}
\end{proposition}

{\bf Proof:} We prove by induction, using an order on all agents that can receive an offer (i.e.,~all agents except for the ones with the color $1$; we will assume that the colors are numbered from $1$ to $NC$). When colors are selected, the step counters of all agents are equal (e.g.,~$sc_i = k$ for all $i$). Agents of the same color have a different $docsId$. Thus, the order between every two agents that can receive an offer is determined first according to their color (small colors come first). If the colors are equal then the tie is broken using their $docsId$ (smaller comes first).


Recall that the conditions for an agent $A_j$ to reply to an offer are that all agents in $PC(j)$ either offered to $A_j$ or their step counter equals $sc_j + 1$. Assume that $A_i$ is the agent with the smallest $docsId$ among the agents with color $2$. It will receive offers from all its neighbors with color $1$. Thus, it will be able to select a neighbor to reply positively to its offer, and all its other neighbors will get a negative reply and unilaterally select an assignment.

The agent with the second smallest $docsId$ that received an offer ($A_j$) with color $2$ can have two types of neighbors with color $1$: Ones that sent an offer to $A_i$ and ones that sent an offer to $A_j$. The ones that sent an offer to $A_i$, after they receive the reply from $A_i$, will attempt to replace their assignment and increase their step counter. After receiving all indications regarding the increase of the step counters of these agents, $A_j$ can reply to the agents that sent it an offer. 

Assume that later on during the algorithm run, $A_i$ is the agent that received an offer, with $sc_i = k$, and with the smallest color index and the smallest $docsId$ among the agents that received an offer and did not yet reply (i.e.,~if agent $A_{i'}$ received an offer and did not yet reply, then either $co_{i'} > co_i$ or $co_{i'} = co_i \& docsId_{i'} > docsId_i$). Since there are no agents with a color smaller than $co_i$ that received an offer and did not reply, then there is no agent that sent an offer with a color index smaller than $co_i - 1$, which a reply was not sent to it. Thus, the members of $PC(i)$ include two types of agents: Agents that sent an offer to $A_i$ and agents that a reply for the offers they sent was already sent to them. Thus, once all the offers from agents of the first type and the indications on the increase in the step counter of the agents from the second type arrive, $A_i$ will be able to reply to the offers sent to it.
$\hfill \square$

\smallskip
An immediate correlation from Proposition~\ref{Prop:cont} is that the algorithm terminates its phases and does not deadlock. The ordering phase uses the DOCS algorithm and its correctness and termination have been established in previous studies~\cite{BarenboimE14,rachmut2022communication}. The pair selection phase must terminate because every agent that receives an offer must reply, and thus, all agents can perform the assignment selection method and increase their step counter.

    
\begin{proposition}
    LAMDLS-2 converges to a $2$-opt solution.
\end{proposition}

{\bf Proof:} According to Proposition~\ref{Prop:mono}, LAMDLS-2 is monotonic. Thus, since the problem is finite, it must converge to some solution. To prove that the solution it converges to is $2$-opt, we need to establish that following convergence, every pair of neighboring agents will get a chance to form a pair and check all their alternative assignments. For agent $A_i$ to form a pair with agent $A_j$, one of them (without loss of generality we select $A_i$) needs to send an offer to the other ($A_j$), and $A_j$ needs to respond positively. This happens in two conditions: (1)~$co_i$ = $co_j -1$; or (2)~for any agent $A_{j'}$ with $co_j = co_{j'}$, $docsId_j < docsId_{j'}$. Since colors and $docsId$s are selected randomly, this situation will eventually occur. 
$\hfill \square$


\section{Extension to a Region-Optimal Algorithm}

Similar to how MGM-2 was extended to $k$-opt and then to region-opt algorithms, we propose an extension of LAMDLS-2 to LAMDLS-ROpt. In LAMDLS-ROpt, an agent initiating ad-hoc coalition formation takes on a mediator role. Unlike LAMDLS-2, where this agent includes its information in the offer message sent to the selected neighbor, in LAMDLS-ROpt, the mediator sends an offer message to neighboring agents within the coalition it aims to form. This message invites them to join and prompts other specified neighbors to join as well. The information of the agents in the forming coalition is sent back to the monitoring agent, who selects an alternative assignment for the group. The group replaces the assignment if the mediator is ordered before the mediators of neighboring groups according to the ordered color and $docsId$ scheme. This process is similar to the region-optimal algorithm RODA~\cite{GrinshpounTLZ19}. The difference is in its repeated selection of mediators, the selection of members in the groups included in the mediators' regions, and the order in which groups replace assignments, in a designated sequence, according to the ordered color scheme used in LAMDLS and LAMDLS-2. We leave for future work the investigation of the performance of LAMDLS-ROpt in comparison with RODA.


\section{Experimental Evaluation} \label{sec:Expirements}
We present a comprehensive study that compares the proposed LAMDLS-2 algorithm to MGM-2, solving a variety of DCOP benchmarks in environments with different patterns of message latency.

\subsection{Experimental Design}
\label{subsec:ED}

In our experiments, we use the same asynchronous simulator used by researchers for CA-DCOP algorithms.\footnote{The simulation's code is available at \url{https://github.com/benrachmut/CADCOP_CP_2024}.} The experiments were conducted on a Windows Server 2019 Standard operating system, with an Intel Xeon Silver 4210 CPU 2.20GHz. 

We follow the approach used in the literature~\cite{rachmut2021latency,rachmut2022communication} to evaluate the quality of the solutions of the algorithms, as a function of the asynchronous advancement of the algorithm, in terms of non-concurrent logic operations (NCLOs)~\cite{ZivanM06b,NetzerGM12}. The utilization of NCLO ensures implementation independence and avoids double counting of simultaneous actions. 

In each experiment, we randomly generated $100$ different problem instances with $50$ agents and we reported the average solution quality of the algorithms examined. To demonstrate the convergence of the algorithms, we present the sum of costs of the constraints involved in the assignment that would have been selected by each algorithm every $10,000$ NCLOs.

We simulated three types of communication scenarios: (1) Perfect communication; (2) Message latency selected from a uniform distribution $U(0,UB)$, where $UB$ is a parameter indicating the maximum latency; and (3) Message latency selected from a Poisson distribution with $\lambda = |MSG|$ and then scaling it by a factor of $m$, where $|MSG|$ represents the number of messages that are currently delivered in the system, and $m$ is a scaling factor indicating the magnitude of the latency. This scenario is the evaluation of the impact of bandwidth load. Latency was also measured in terms of NCLOs.

We evaluated our algorithms on three problem types that are commonly used in the DCOP literature: 
\begin{itemize}

\item \textbf{Uniform Random Problems}. These are random constraint graph topologies with densities $0.2$ and $0.7$. Each variable had a domain of $10$ values, and constraint costs were uniformly selected between $1$ and $100$. 
\item \textbf{Graph Coloring Problems}~\cite{zhang:05,farinelli:08}. Each variable has three values (colors). Equal assignments between two neighbors incurred random costs from $U(10,100)$, while non-equal assignments had $0$ cost. The density was set at $0.05$. 
\item \textbf{Scale free Network Problems}~\cite{barabasi:99}. Initially, $10$ agents were randomly selected and connected. Additional agents were sequentially added, connecting to $3$ other agents with probabilities proportional to the existing agents' edge counts. Similar to the first type, variables had a domain of $10$ values, and constraint costs ranged from $1$ to $100$. 
\end{itemize}

\subsection{Experimental Evaluation}
\label{subsec:EE}

\begin{figure}[t]
\vspace{-8pt}
 \begin{center}
 \hspace{-10pt}
 \begin{tabular}{cc}
	\includegraphics[scale=0.40]{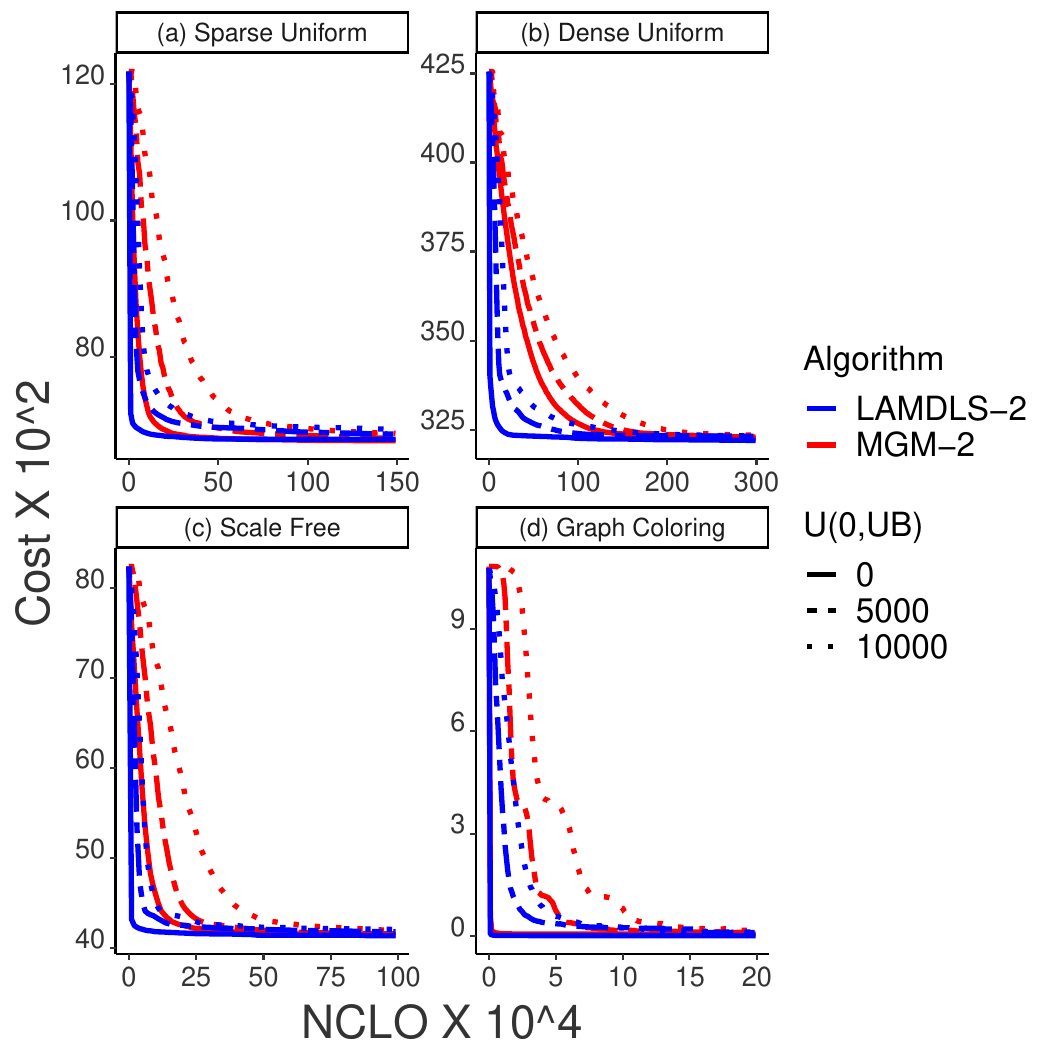} &
		
 \end{tabular}
\vspace{-5pt}
 \caption{ Solution quality as a function of NCLOs. Message delays are sampled from a uniform distribution.}
 
 \label{Fig:uniform_delay}
 \vspace{-20pt}
 \end{center}
\end{figure}

\begin{figure}[!t]
 \begin{center}
 \hspace{-10pt}
 \begin{tabular}{cc}
	\includegraphics[scale=0.40]{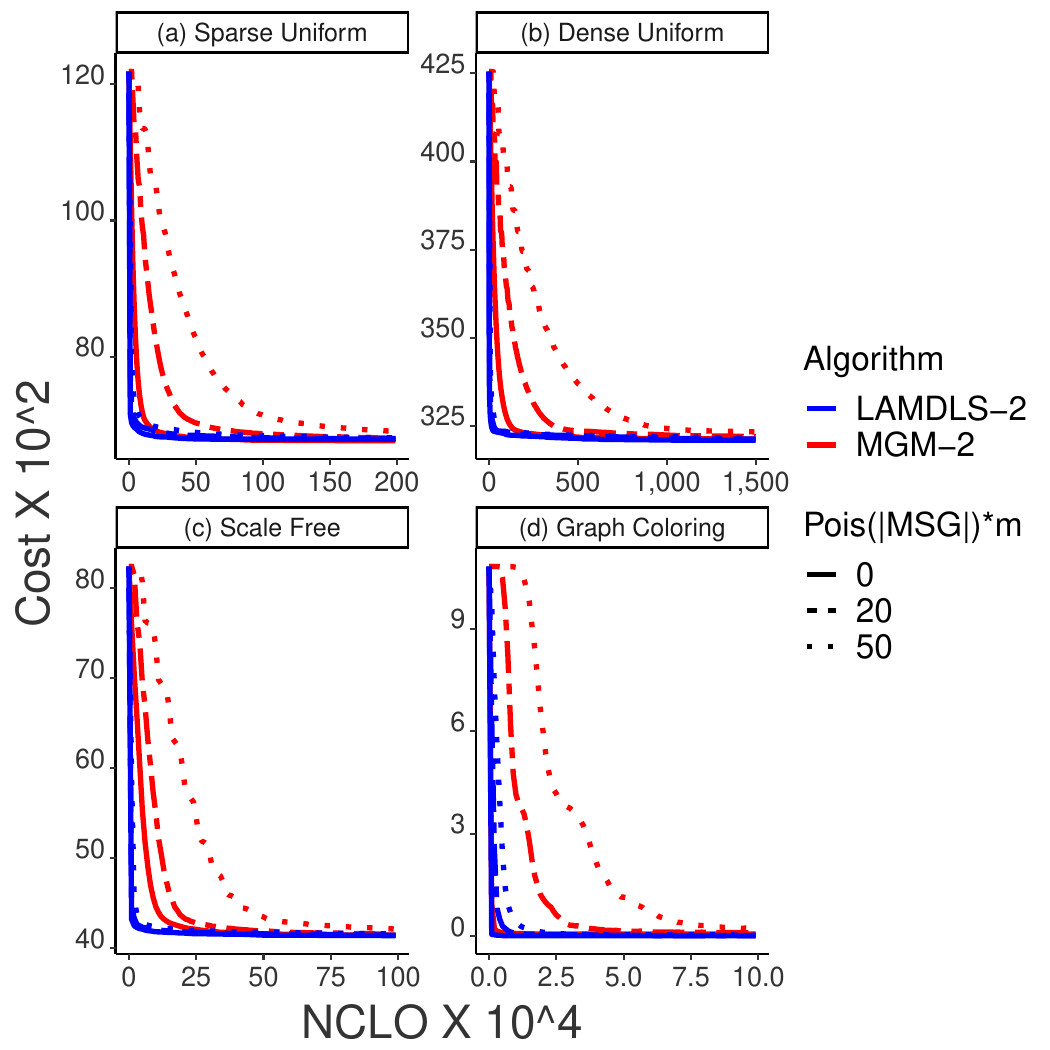} &
 \end{tabular}
\vspace{-5pt}
 \caption{Solution quality as a function of NCLOs. Message delays sampled from a Poisson distribution linked to message volume.} 
 \label{Fig:msg_delay}
 \vspace{-20pt}
 \end{center}
\end{figure}

Figure~\ref{Fig:uniform_delay} presents a comparison between the results of two algorithms: The proposed LAMDLS-2 (represented by the blue curve) and MGM-2 (represented by the red curve). The comparison is performed on different problem types, as shown in each subgraph. The graph illustrates the performance of both algorithms in terms of the average global cost as a function of NCLOs. This enables the demonstration of the solution quality and the convergence speed for each algorithm. Latency is sampled from a uniform distribution, and the line type (solid, dashed, and dotted) corresponds to different magnitudes of latency, where $UB = \{0$, $5,\!000$, $10,\!000\}$. The results demonstrate that the algorithms converge to solutions with similar quality, independent of message delays. This is expected because, in both algorithms, agents wait for updated information from their neighbors before they perform computation and replace assignments.

LAMDLS-2 demonstrates faster convergence than MGM-2 in scenarios with no message delays, except when solving graph coloring problems, where both algorithms show similar convergence rates. Moreover, LAMDLS-2 is more resilient to message delays than MGM-2. Its convergence rate remains relatively stable even with increasing delay, while MGM-2 experiences a more substantial slowdown in convergence as the latency magnitude increases. The most significant difference in the convergence rate between LAMDLS-2 and MGM-2 is observed in dense uniform problems (Figure~\ref{Fig:uniform_delay}(b)). Interestingly, LAMDLS-2 with the longest delays $UB = 10,\!000$ converges faster than MGM-2 with no delays. When solving graph coloring problems (Figure~\ref{Fig:uniform_delay}(d)), although the convergence rates are similar when communication is perfect, LAMDLS-2 exhibits a much faster convergence rate compared to MGM-2 when messages are delayed. These problems are characterized by low density among the examined types, leading to rapid convergence for both algorithms. For sparse uniform problems (Figure~\ref{Fig:uniform_delay}(a)), the impact of message delays on both LAMDLS-2 and MGM-2 is consistent and proportional. However, LAMDLS-2 maintains its superiority over MGM-2 in terms of convergence speed. When solving scale-free networks (Figure~\ref{Fig:uniform_delay}(c)), the negative impact on convergence rates is more pronounced for MGM-2 compared to LAMDLS-2 as the latency magnitude increases. Figure~\ref{Fig:msg_delay} presents the results of a similar experiment in which message delays were sampled from a Poisson distribution with the parameter $\lambda = |MSG| \cdot m$, where $m = \{0$, $20$, $50\}$. In this set of experiments, the resilience of LAMDLS-2 is pronounced regardless of the type of problem being solved. The increase in the latency magnitude did not significantly affect LAMDLS-2's convergence rate, unlike the significant effect it had on MGM-2.

\begin{figure}[!t]
 \begin{center}
 \hspace{-10pt}
 \begin{tabular}{cc}
	\includegraphics[scale=0.40]{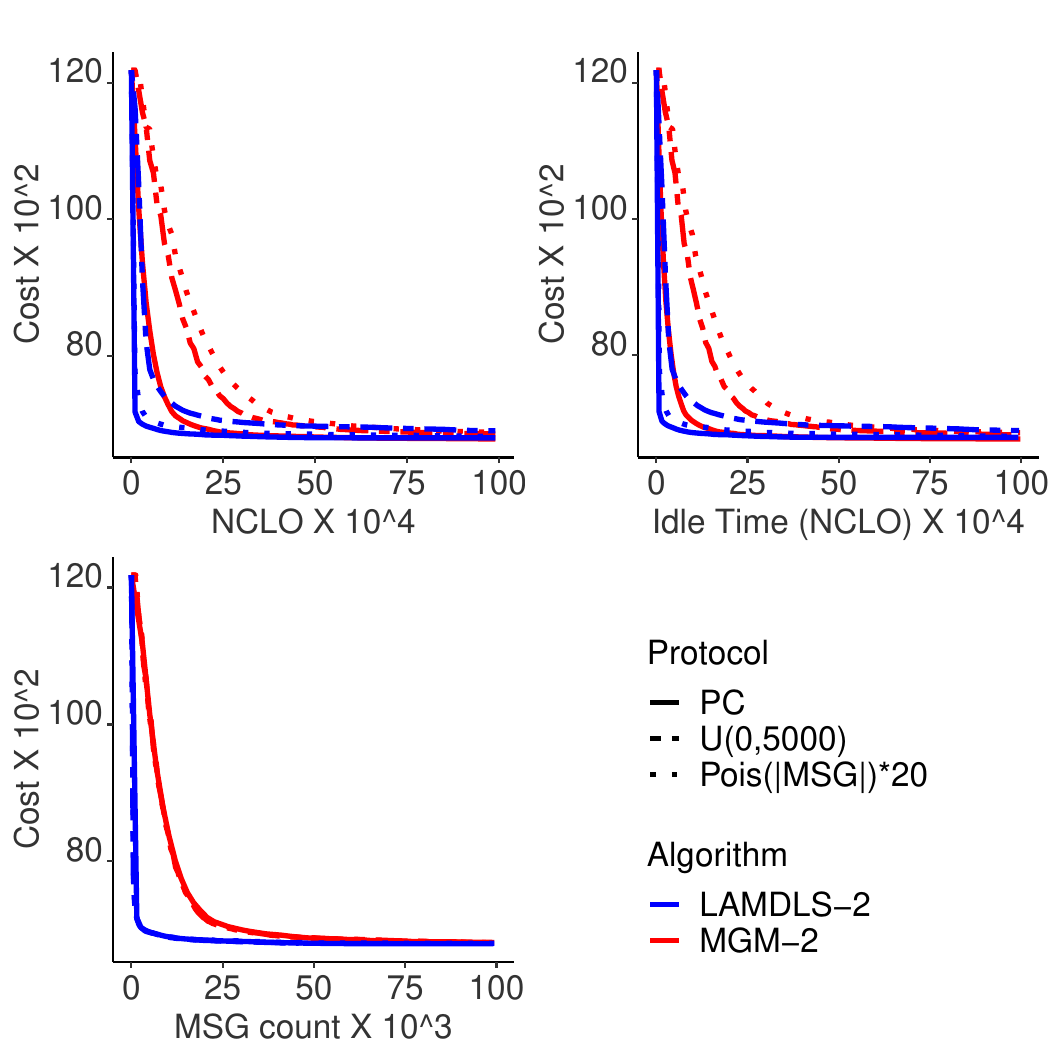} &
 \end{tabular}
\vspace{-5pt}
 \caption {Solution quality as a function of different matrices in environments with different message delays. } 
 \label{Fig:measures}
 \vspace{-20pt}
 \end{center}
\end{figure}

The results in Figures~\ref{Fig:uniform_delay} and~\ref{Fig:msg_delay} indicate a faster convergence rate of LAMDLS-2 in comparison with MGM-2. To investigate the reasons for this advantage, we present in Figure~\ref{Fig:measures} the solution costs of the algorithms as a function of two additional elements in the algorithms' execution. These elements are the number of messages exchanged by the agents and the amount of time (in NCLOs) that agents were inactive (i.e., idle). Both algorithms solve sparse uniform problems under various communication scenarios: Perfect communication (PC) represented by the solid line, $U(0$,$5,\!000)$ represented by the dashed line, and $Pois(|MSG|)\cdot20$ represented by the dotted line. While the three presented subgraphs illustrate the faster convergence rate of LAMDLS-2 compared to MGM-2, each of them highlights a distinct advantage of LAMDLS-2. The faster convergence in terms of message count indicates that LAMDLS-2 makes more economical use of the communication network. The faster convergence in terms of idle time indicates that agents in LAMDLS-2 are more active, and perform more concurrently. 

\begin{figure}[!t]
 \begin{center}
 \hspace{-10pt}
 \begin{tabular}{cc}
	\includegraphics[scale=0.40]{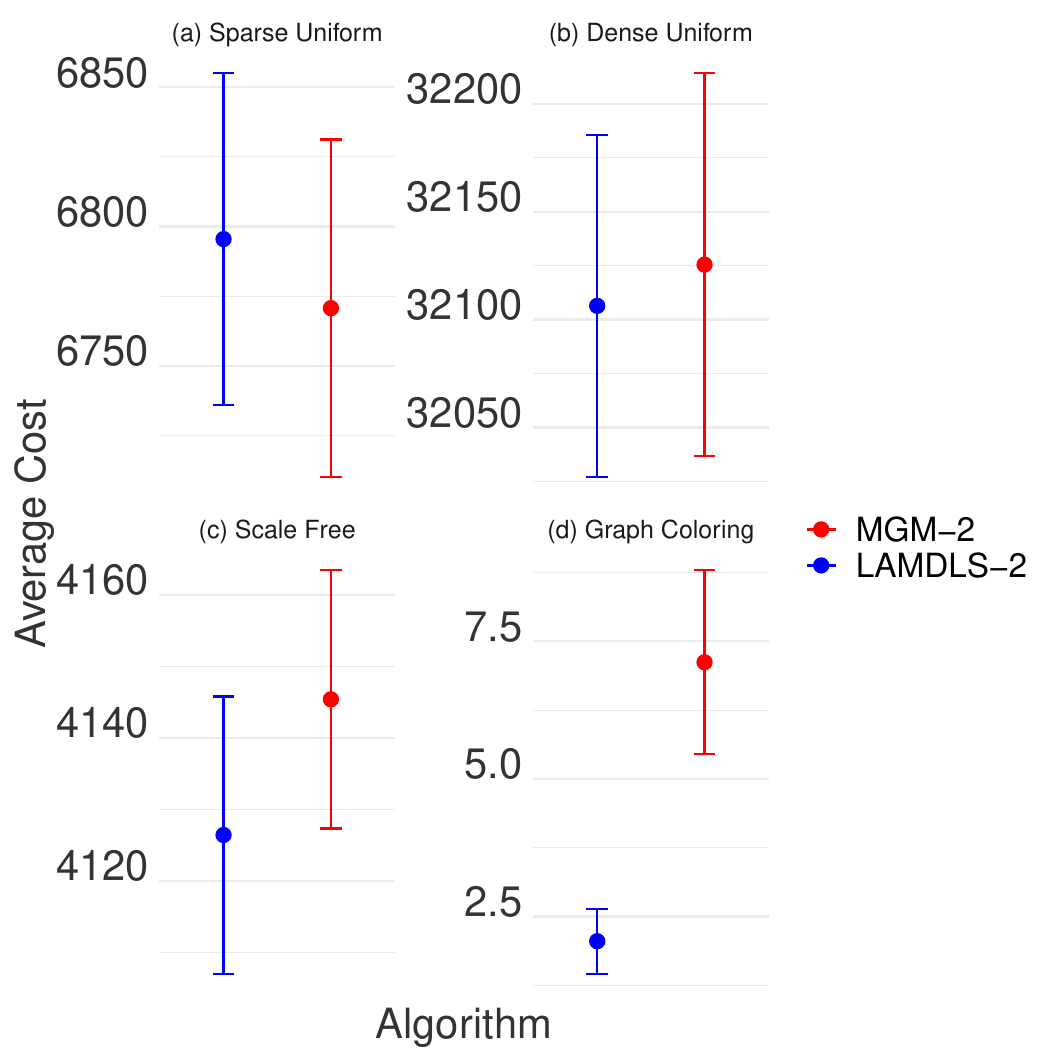} &
 \end{tabular}
\vspace{-5pt}
 \caption {\centering Average costs at convergence with error bars.}
 \label{Fig:stat}
 \vspace{-20pt}
 \end{center}
\end{figure}

In Figure~\ref{Fig:stat}, we present the average costs of both algorithms at convergence with SEM error bars. Overlapping bars across sparse, dense, and scale-free networks suggest no significant difference. Paired t-tests confirm this, with p-values above 0.05 ($0.7514$ for sparse, $0.8364$ for dense, and $0.4839$ for scale-free). For graph coloring problems, there is a significant difference (p-value $0.005$), indicating diverse algorithmic performance in favor of LAMDLS-2.

\section{Conclusions}

We introduced Latency-Aware Monotonic Distributed Local Search 2 (LAMDLS-2), a distributed local search algorithm for solving DCOPs, which is monotonic and guarantees convergence to a $2$-opt solution. LAMDLS-2 converges faster, compared to MGM-2, a synchronous distributed local search algorithm that converges to $2$-opt solutions with similar quality. We demonstrate that the algorithm not only converges faster but also makes more economical use of the communication network and that the agents spend less time idle during the algorithm run. The results indicate that LAMDLS-2 is more suitable for realistic scenarios with message delays. Our approach, which is based on the ordered color scheme, allows the agents to be more active in computing their assignments and spend less effort in coordinating their actions. We also discussed how this approach can be extended to a general $k$-opt algorithm, which we intend to implement in future work.

\bibliography{LAMDLS}

\end{document}

%% file: commends.tex
\usepackage{algorithm}
\usepackage{algpseudocode}
\usepackage{comment} 

\newcommand{\algorithmicinput}{\textbf{Input:}}
\newcommand{\algorithmicoutput}{\textbf{Output:}}
\newcommand{\algorithmicinputtext}{\item[\algorithmicinput]}
\newcommand{\algorithmicoutputtext}{\item[\algorithmicoutput]}

\usepackage[table,xcdraw]{xcolor}
\newcommand{\ben}[1]{\textcolor{black}{#1}}

\newcommand{\benLine}[1]{\textcolor{black}{#1}}

\usepackage{todonotes}
\usepackage{ifthen}
\newboolean{includeMemo}
\setboolean{includeMemo}{false}

\newcommand{\memoside}[1]{\ifthenelse{\boolean{includeMemo}}{\todo[caption={},color=green!20!]{{\footnotesize #1}}}{\relax}}
\newcommand{\memo}[1]{\ifthenelse{\boolean{includeMemo}}{\todo[inline,caption={},color=green!20!]{#1}}{\relax}}
\usepackage[toc,page]{appendix}